\documentclass[lettersize,journal]{IEEEtran}
\usepackage{amsmath,amsfonts}
\usepackage{algorithmic}
\usepackage{algorithm}
\usepackage{array}
\usepackage[caption=false,font=normalsize,labelfont=sf,textfont=sf]{subfig}
\usepackage{textcomp}
\usepackage{stfloats}
\usepackage{url}
\usepackage{multirow}
\usepackage{verbatim}
\usepackage{graphicx}
\usepackage{cite}
\usepackage{color}
\usepackage{makecell}
\begin{document}
\title{Unsupervised Domain Adaptation on \\ Person Re-Identification via \\ Dual-level Asymmetric Mutual Learning}

\author{
Qiong Wu,
Jiahan Li,
Pingyang Dai,
Qixiang Ye,~\IEEEmembership{Senior Member,~IEEE},
Liujuan Cao,~\IEEEmembership{Member,~IEEE}, \\
Yongjian Wu,
Rongrong Ji,~\IEEEmembership{Senior Member,~IEEE}

\thanks{Qiong Wu is with Institute of Artificial Intelligence, and the Media Analytics and Computing Laboratory, Department of Artificial Intelligence, School of Informatics, Xiamen University, Xiamen 361005, China (e-mail: qiong@stu.xmu.edu.cn).}

\thanks{Jiahan Li is with School of Information and Control Engineering, China University of Mining and Technology, Xuzhou 221000, China (e-mail: jiahan.li@cumt.edu.cn).}

\thanks{Pingyang Dai is with the Media Analytics and Computing Laboratory, Department of Artificial Intelligence, School of Informatics, Xiamen University, Xiamen 361005, China (e-mail: pydai@xmu.edu.cn).}

\thanks{Qixiang Ye is with the Peng Cheng Laboratory, Shenzhen 518066, China, and also with the School of Electronics, Electrical and Communication Engineering, University of Chinese Academy of Sciences, Beijing 100049, China (e-mail: qxye@ucas.ac.cn).}

\thanks{Liujuan Cao is with the Media Analytics and Computing Lab, Department of Computer Science, School of Informatics, Xiamen University, Xiamen 361005, China (e-mail: caoliujuan@xmu.edu.cn).}

\thanks{Yongjian Wu is with the Youtu Laboratory, Tencent, Shanghai 200233,
China. (e-mail: littlekenwu@tencent.com).}

\thanks{Rongrong Ji is with the Media Analytics and Computing Laboratory, Department of Artificial Intelligence, School of Informatics, Xiamen University, Xiamen 361005, China, also with the Fujian Engineering Research Center of Trusted Artificial Intelligence Analysis and Application, Institute of Artificial Intelligence, Xiamen University, Xiamen 361005, China, and also with the Peng Cheng Laboratory, Shenzhen 518066, China. (e-mail: rrji@xmu.edu.cn).}
}

\maketitle

\markboth{Journal of \LaTeX\ Class Files,~Vol.~14, No.~8, August~2021}%
{Shell \MakeLowercase{\textit{et al.}}: A Sample Article Using IEEEtran.cls for IEEE Journals}


\begin{abstract}
Unsupervised domain adaptation person re-identification (Re-ID) aims to identify pedestrian images within an unlabeled target domain with an auxiliary labeled source-domain dataset.
Many existing works attempt to recover reliable identity information by considering multiple homogeneous networks.
And take these generated labels to train the model in the target domain.
However, these homogeneous networks identify people in approximate subspaces and equally exchange their knowledge with others or their mean net to improve their ability, inevitably limiting the scope of available knowledge and putting them into the same mistake.
This paper proposes a Dual-level Asymmetric Mutual Learning method (DAML) to learn discriminative representations from a broader knowledge scope with diverse embedding spaces.
Specifically, two heterogeneous networks mutually learn knowledge from asymmetric subspaces through the pseudo label generation in a hard distillation manner.
The knowledge transfer between two networks is based on an asymmetric mutual learning manner.
The teacher network learns to identify both the target and source domain while adapting to the target domain distribution based on the knowledge of the student.
Meanwhile, the student network is trained on the target dataset and employs the ground-truth label through the knowledge of the teacher.
Extensive experiments in \textbf{Market-1501}, \textbf{CUHK-SYSU}, and \textbf{MSMT17} public datasets verified the superiority of DAML over state-of-the-arts.
\end{abstract}


\vspace{2mm}
\section{Introduction}

\begin{figure}[t]
\centering
\includegraphics[width=1.0\columnwidth]{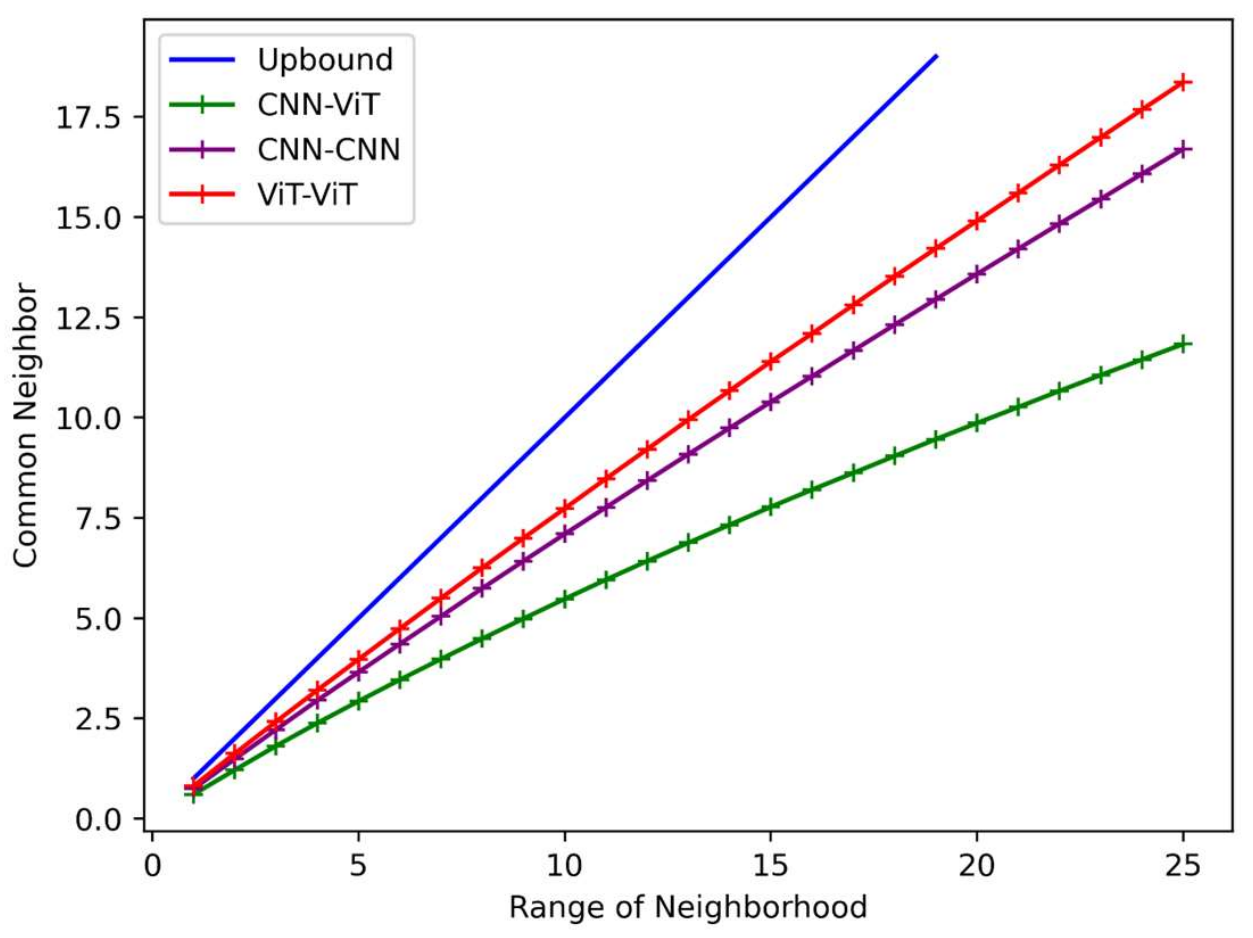}
\caption{\small
The statistics of common neighbors between different models.
CNN-CNN and ViT-ViT curves denote the average number of common neighbors in the k nearest neighbors of each instance.
The features are extracted by two homogeneous networks trained with different initialization.
Similarly, CNN-ViT represents the common neighbors between CNN and ViT.
Furthermore, Upbound refers to the maximum number of neighbors to consider.
These models are trained on the CUHK-SYSU dataset in a supervised manner and cluster on the Market1501 dataset.
Compared to the CNN-CNN and ViT-ViT, the CNN-ViT contains fewer common neighbors, and the ways they distinguish two individuals are more different than homogeneous networks.
It demonstrates that heterogeneous networks address the task in different patterns.
}
\label{fig1:Motivation}
\end{figure}

\IEEEPARstart{P}{erson} re-identification (Re-ID)~\cite{series/acvpr/GongCLH14} aims at matching individual pedestrian images from images captured by different cameras according to identity.
\IEEEpubidadjcol
This task is challenging because the variations of viewpoints, body poses, illuminations, and backgrounds will influence a person's appearance.
Recently, supervised person Re-ID methods~\cite{journals/ijcv/LiZG20, journals/pami/SunZLYTW21, journals/ijcv/YinWZ20, conf/eccv/SunZYTW18, conf/cvpr/ZhangLZJ020, conf/cvpr/WangZGL18, conf/iccv/WuZL19, conf/cvpr/YuZ20, conf/eccv/ZhongZLY18, conf/iccv/HanYZTZGS19} made impressive progress.
However, as the number of images increases and the ensuing scene changes, regular supervised learning approaches are losing their ability to adapt to complex scenarios.
The performance of person re-ID models trained on existing datasets will evidently suffer for person images from a new video surveillance system due to the domain gap.
%
%
%
To avoid time-consuming annotations on the new dataset, unsupervised domain adaptation (UDA) is proposed to adapt the model trained on the labeled source-domain dataset to the unlabeled target-domain dataset.
%

Generating trusted identity information on the target domain is seen as the core of the UDA task.
Some UDA Re-ID methods~\cite{conf/cvpr/Deng0YK0J18, conf/iccv/LiLLW19, conf/cvpr/WeiZ0018, conf/eccv/ZhongZLY18, conf/eccv/ZouYYKK20} directly apply GANs~\cite{journals/corr/GoodfellowPMXWOCB14} to transfer the style of pedestrian images from the source domain to the target while keeping the identities to train the model.
However, the complexity of the human form and the limited number of instance in a Re-ID dataset limit the quality of generated images.
%
%
%
After abandoning the image generation, some methods~\cite{conf/aaai/ChangYXH19, conf/cvpr/WangZGL18} introduce the attribute to bridge the domain gap.
These methods introduce additional annotation information which defeats the purpose of the UDA Re-ID task.
Limited by the missing label on the target domain, others~\cite{conf/iccv/QiWHZSG19, conf/cvpr/Zhong0LL019, conf/cvpr/YangZLCLLS21, journals/tip/DaiCWHYTLJ22} align the distributions of target and source domains while only learning classifying on the source.
To better adapt the distribution of the target domain and train with the target-domain identity knowledge, various methods~\cite{conf/aaai/LinD00019, conf/iccv/ZhangCSY19, conf/cvpr/BaiWW0D21} apply a clustering algorithm in the target domain to generate the pseudo labels for training in a supervised manner. 
One of the keys to improving performance is alleviating the influence of noisy labels. 
%
In this context, many methods~\cite{conf/iclr/GeCL20, conf/cvpr/ZhengLH0LZ21, conf/eccv/ZhaiYLJJ020} based on clustering algorithms are proposed to rule out the harmony from the noisy labels by introducing more than one framework to predict pseudo labels.
%
%
%
%
They aim to generate knowledge with specific differences in samples and exchange the knowledge among the networks to enhance their ability.
%
Despite encouraging progress, the benefits from the knowledge mined by homogeneous networks are limited.
As shown in Fig.~\ref{fig1:Motivation} CNN-CNN and ViT-ViT, these homogeneous networks with similar structures identify pedestrians in a comparable manner, and the relation among the instances are similar.
%
It suggests they use similar patterns to extract pedestrian features, and networks may converge to equal each other.
Furthermore, this mode of operation makes it possible for the networks to make the same mistakes and not be able to correct them.
Such a design limits the knowledge models can learn from the training set and makes it possible for the networks to repeat mistakes without being able to remedy them.
As a result, mining the information from different subspaces is required to broaden the scope of knowledge and generate reliable pseudo labels.

To tackle this problem, heterogeneous networks, as shown in Fig.~\ref{fig1:Motivation} CNN-ViT, can discover the information from multiple subspaces and have more extensive latent knowledge.
%
%
We propose Dual-level Asymmetric Mutual Learning (DAML), a novel unsupervised domain adaption method for person Re-ID that broadens the scope of knowledge for the network by exploiting information from two different subspaces and selectively transferring information between heterogeneous networks.
The proposed DAML consists of a CNN that focuses on identity learning as a teacher network and a ViT that concentrates on adapting knowledge from the target domain as a student network for embedding samples into different subspaces and setting the constraints among the classifiers for asymmetric mutual learning.

In particular, the CNN that works as a teacher will train on both source and target datasets under the supervision of ground-truth source-domain labels and pseudo-target-domain labels.
The former can provide reliable identity information for extracting discriminative feature representation, while the latter will assist the network in adapting the  distribution of the target domain.
However, learning from the source domain will harm the distribution that the network adapted limiting the performance.
To avoid this disadvantage, the ViT that works as a student only trains with the guidance of pseudo-target-domain labels and learns the knowledge from the teacher.
%
%
%
%
%
%
In the pseudo label generation stage, the relationship between two samples is weighted according to their teacher and student features similarity.
Moreover, this process wholly exchanges the knowledge learned from two different subspaces.
%
After predicting the identities of input images, the asymmetric constraints between two heterogeneous networks selectively exchange the knowledge.
The student learns the identity knowledge from the teacher network under the constraints from the target-domain samples.
Furthermore, for the student can benefit more from the teacher and better utilize the ground-truth labels, the source-domain identity knowledge learned by the teacher is transferred to the target domain with the constraints based on source-domain samples.
%
In summary, the DAML employs diverse subspaces to generate reliable pseudo label in the target domain and help student adopt ground-truth knowledge in the source domain.

Our main contributions are summarized below:

\begin{itemize}
    \item We address the diverse subspaces learning and target-domain identity learning for unsupervised domain adaptation person Re-ID with proposed Dual-level Asymmetric Mutual Learning (DAML). 
    The former has rarely been studied in the existing research, while the latter is crucial for retrieving person in the target domain.
    \item We propose a novel Dual-level Asymmetric Mutual Learning (DAML) method for unsupervised domain adaptation person Re-ID. 
    The asymmetric knowledge learning between the teacher and the student helps them play their roles better.
    \item To learn from diverse subspaces, the proposed DAML introduces two heterogeneous networks to mine valuable information from different subspaces and selectively exchange the information between them.
    \item To better utilize the knowledge mined by heterogeneous networks and ensure the networks orient to the task, the proposed DAML smoothly update the classifiers in a hard distillation manner and exchange knowledge during training in a soft distillation manner.
\end{itemize}

\section{Related Work}

\subsection{Unsupervised Domain Adaptation Person Re-ID}
Unsupervised Domain Adaptation Person Re-ID has attracted increasing attention in recent years due to its effectiveness in reducing manual annotation costs.
There are two main categories of methods are proposed to address this issue.
Firstly, GAN-based methods aim to transfer samples from the source domain to the target domain without altering their identities. 
SPGAN~\cite{conf/cvpr/Deng0YK0J18} and PDA-Net~\cite{conf/iccv/LiLLW19} transfer images directly from the source domain to the target domain while maintaining the original identity knowledge.
The generated images have a similar style to the target-domain images and are used to train the model under the supervision of their original labels in the source domain.
To produce generated images that are more realistic and have more detail, DG-Net~\cite{conf/cvpr/ZhengYY00K19} and DG-Net++~\cite{conf/eccv/ZouYYKK20} introduce disentanglement for the generation stage.
But the generation is expensive and the style of generated images may not well fit the target domain.
Rather than transfer images from the source domain to the target domain, HHL~\cite{conf/eccv/ZhongZLY18} transfers target-domain images among the cameras to generate images that have the same identity but contain the difference at the same time.
%
%
Secondly, the clustering-based methods clustering based methods do not require expensive GAN networks for generation and have achieved state-of-the-art performance to date.
To reduce the impact of noisy label, MMT~\cite{conf/iclr/GeCL20} proposed a mutual learning method providing soft labels.
For more reliable pseudo labels, SSG~\cite{conf/iccv/FuWWZSUH19} clusters samples in three scales and validate each other.
MEB-Net~\cite{conf/eccv/ZhaiYLJJ020} respectively introduces multiple groups of prototypes or homogeneous networks to generate the pseudo labels.
UNRN~\cite{conf/aaai/ZhengLZZZ21} and GLT~\cite{conf/cvpr/ZhengLH0LZ21} design a memory bank to save anchors for aligning the distribution and learning identities in a contrastive learning manner.
Limited by the constraints in the feature level these methods rely on, the models that collaborate to generate pseudo labels are homogeneous.
These characteristics determine that the model can only learn similar knowledge from others.
Nevertheless, these approaches alleviate the domain gap only considering the single embedding space inevitably makes some mistakes.

\subsection{Knowledge Distillation}
Knowledge distillation makes a student network learns from a strong teacher network to improve the student's ability.
The common approaches can be summarized as hard distillation and soft distillation.
Soft distillation~\cite{journals/corr/HintonVD15, conf/eccv/WeiXXZ0T20} minimizes the distribution difference between the prediction generated by teacher and student.
The soft label generated by the teacher model can alleviate overfitting just like labels smoothing~\cite{conf/cvpr/YuanTLWF20}.
Unlike the soft, hard-label distillation regards the prediction result of the teacher as a valid label.
And positive pairs predicted by the teacher are used to transfer identity knowledge from the teacher to a student network in semi-supervised and unsupervised learning tasks.
Temporal ensembling~\cite{conf/iclr/LaineA17} put the former networks as the teacher and use memory saving average predictions for each sample as supervision for the unlabeled samples.
To avoid storing predictions for saving memory, Mean Teacher~\cite{conf/iclr/TarvainenV17} averaged student model weights as the parameter of the teacher.
During the training, the predictions made by the teacher are seen as supervision for unlabeled samples.
The models consider similar information in these methods because the teacher and the student have the same structure and similar initialization.
It makes the networks focus within a certain range and limit the knowledge student can learn.
The proposed DAML exchange knowledge utilizes both soft and hard distillation in the different training stages.
Thanks to the heterogeneous networks, the proposed DAML gives the student model a broader perspective and can generate pseudo labels from different views.

\begin{figure*}[t]
\centering
\includegraphics[width=0.9\textwidth]{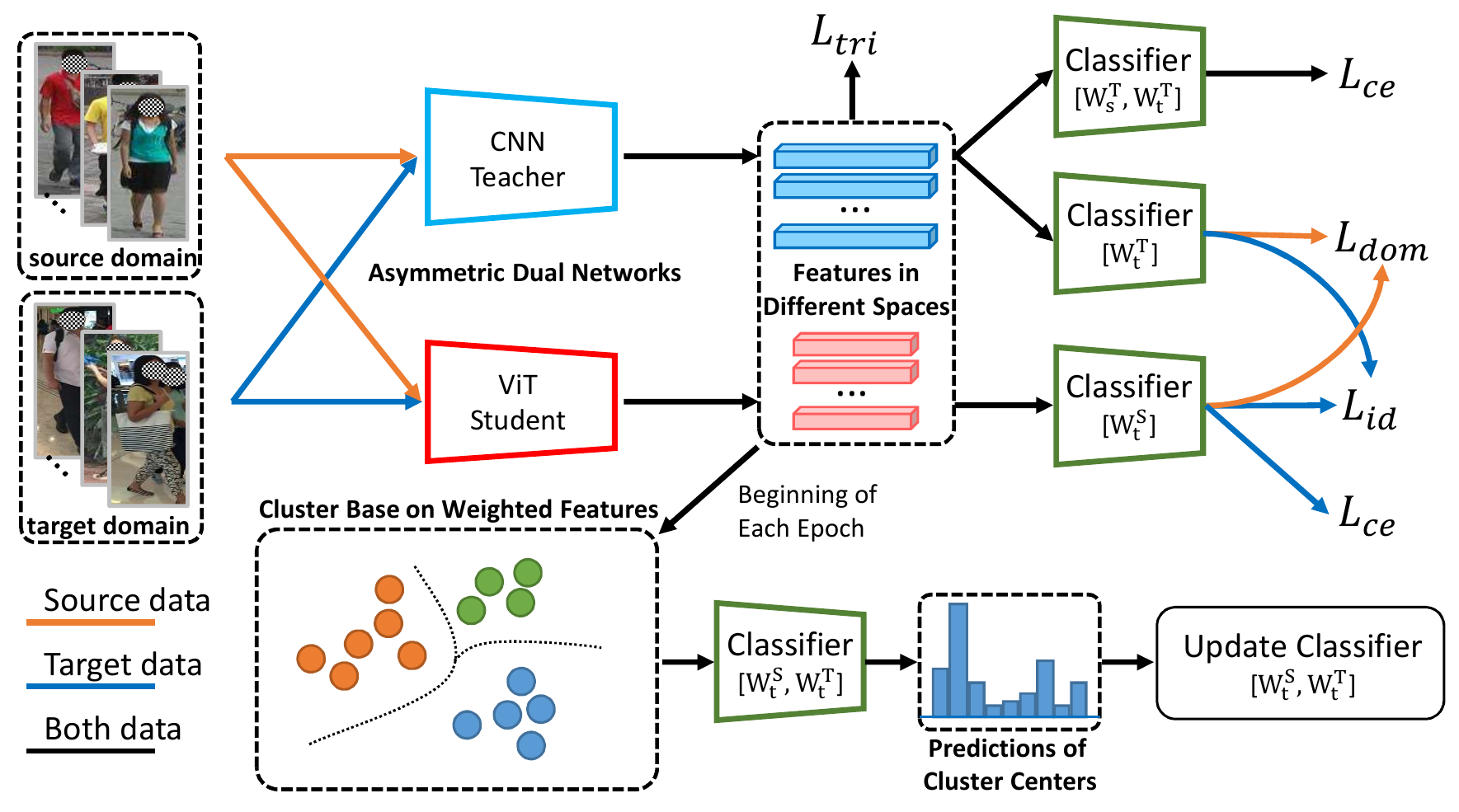}
\vspace{-3mm}
\caption{ \small
Overview of our Dual-level Asymmetric Mutual Learning method (DAML).
%
The teacher network is trained under the supervision of pseudo labels and ground-truth labels for target-domain and source-domain samples.
And the student only directly learns knowledge from target-domain samples with pseudo labels.
At the beginning of epochs, we first generate the pseudo labels for target dataset, and update the classifiers based on the predictions of cluster centers.
To distill the different subspace knowledge from the teacher to the student,  $\mathcal{L}_{id}$ makes the student predictions of target-domain samples close to the teacher.
%
Meanwhile, for student can better adopt the identity knowledge learned by the teacher, we minimize the distribution differences of the same source-domain samples with $\mathcal{L}_{dom}$.
}
\label{fig2:Framework}
\end{figure*}

\subsection{CNN and ViT}
Since AlexNet~\cite{conf/nips/KrizhevskySH12} achieve great success on ImageNet~\cite{conf/cvpr/DengDSLL009}, a variety of convolutional neural networks (CNN)~\cite{conf/cvpr/SzegedyVISW16, conf/cvpr/HeZRS16, conf/cvpr/HuangLMW17, conf/eccv/PanLST18} is proposed to solve different tasks.
As Transformers~\cite{conf/nips/VaswaniSPUJGKP17} were proposed for machine translation and were seen with significant results in many NLP tasks, the application of self-attention to images is widely concerned. 
A new model without any convolution, Vision Transformers (ViT)~\cite{conf/iclr/DosovitskiyB0WZ21}, has been proposed for computer vision tasks and shows its potential.
During the calculation process, the CNN keeps the spatial information and can only focus on the surrounding area in one layer due to the nature of convolution.
In contrast, ViT emphasizes the correlation between two patches, and its receptive field involves the whole feature map.
These differences make the CNN and ViT learn different knowledge from the training set for the same task.
And in our paper, we take advantage of this difference to achieve asymmetric distillation, making ViT a better performer with our DAML.
The ViT works as a student because the receptive field of a patch in the ViT covers the area that one convolution kernel can consider, not vice-versa. 

\section{Methodology}
\label{Sec:Methodology}
\subsection{Overview}

The ultimate goal of the unsupervised domain adaptation (UDA) person Re-ID is to gain a model work on a target-domain dataset based on a labeled source-domain dataset and an unlabeled target-domain dataset.
Let $\mathcal{S}=\{(\mathbf{x}_s^i, \mathbf{y}_s^i)\}_{i=1}^{N_s}$ and $\mathcal{T}=\{\mathbf{x}_t^i\}_{i=1}^{N_t}$ respectively denote the source-domain images with ground-truth labels and the unlabeled target-domain images, where $N_s$ and $N_t$ are the numbers of samples from these two domains.

As shown in Fig.~\ref{fig2:Framework}, the Dual-level Asymmetric Mutual Learning (DAML) method trains the student to extract discriminative representations from two different subspaces to perform the UDA person Re-ID task.
Firstly, DAML adopts two heterogeneous networks: teacher CNN ${\rm E}_T(\cdot)$ and student ViT ${\rm E}_S(\cdot)$ which are pre-trained on the source-domain dataset in a supervised manner to extract features in different subspaces.
At each epoch, we first group target-domain samples into $K$ classes by the clustering algorithm.
The distance between two target-domain samples will be calculated according to the features ${\rm E}_T(\mathbf{x}_t^i)=\mathbf{t}_i^T \in \mathbb{R}^{c_T}$ and ${\rm E}_S(\mathbf{x}_t^i)=\mathbf{t}_i^S \in \mathbb{R}^{c_S}$ extracted by the teacher and student models with corresponding weights.
The $\{\hat{\mathbf{y}}_i\}_{i=1}^{N_t}$ are the pseudo labels for the target-domain samples.
Then, for each class center $\mathbf{c}_y$, we generate its prediction with the classifiers ${\rm C}(\cdot|\mathbf{W}_t^S)$ and ${\rm C}(\cdot|\mathbf{W}_t^T)$ for updating the parameter $\mathbf{W}_t^S$ and $\mathbf{W}_t^T$ in a smooth method.

After that, we train the teacher and the student models with the pseudo labels in a supervised manner.
For the teacher model, classifier ${\rm C}(\cdot|[\mathbf{W}_s^T, \mathbf{W}_t^T])$ will learn both source-domain and target-domain knowledge.
While the classifier ${\rm C}(\cdot|\mathbf{W}_t^S)$ for the student model only directly learns the target-domain knowledge.
%
%
%
The constraints between two networks transfer the identity knowledge learned by the teacher to the target and help the student learn from diverse subspaces. 
%
%
Finally, we only adopt the features $\mathbf{t}_i^S = {\rm E}_S(\mathbf{x}_t^i)$ extracted by the student model for testing.

\subsection{Smooth Classifier Update (SCU)}

At the beginning of epochs, we extract the target-domain features $\mathbf{t}_i^T = E_T(\mathbf{x}^i_t)$ and $\mathbf{t}_i^S = E_S(\mathbf{x}^i_t)$ with two heterogeneous networks.
To better utilize the knowledge from the two models, we first define the neighborhood of an instance according to its relations in two different subspaces:
\begin{equation}
    N_i = \Bigg\{x_j \Bigg| 1 - \frac{\langle t_i^M, t_j^M \rangle}{\Vert t_i^M\Vert_2 \Vert t_j^M\Vert_2} < \alpha, M \in \{T, S\} \Bigg\},
\end{equation}
the $\alpha$ here is a hyper-parameter.
With the limitation of neighbor selection considering both teacher features and student features simultaneously, the neighbors of an instance should be close to it in both subspaces.
The above constraint ensures that instances with apparent differences will not be clustered as the same identity because the patterns of the two models applied to recognize an instance are different.

To exploit the information from two different subspaces and make the pseudo labels more reliable, we combine features from heterogeneous networks and define the distance between two samples as:
\begin{equation}
    d_{i,j} =
\begin{cases}
\begin{aligned}
& 1 - \frac{\langle[\mathbf{t}_i^T, \mathbf{t}_i^S],[\mathbf{t}_j^T, \mathbf{t}_j^S]\rangle}
           {\Vert[\mathbf{t}_i^T, \mathbf{t}_i^S]\Vert_2 \Vert[\mathbf{t}_j^T, \mathbf{t}_j^S]\Vert_2}, & x_i \in N_j \ and \ x_j \in N_i \\
& \rm{Inf}, & Others
\end{aligned}
\end{cases}
\end{equation}
where $[\cdot,\cdot]$ represents the concatenation of two features, and $\langle\cdot,\cdot\rangle$ denotes the dot product between two features.
In short, we define the similarity between two samples as the cos similarity between the features constructed by concatenating their teacher feature and student feature.
Then the pseudo labels can be generated based on the relationship among instances with the clustering algorithm.
With the pseudo labels, some methods \cite{conf/iclr/GeCL20,conf/aaai/ZhengLZZZ21} directly update the classifier by replacing the classifier parameters with the new class centers to adapt the count of classes change.
These methods will make the knowledge lost because the class centers may not represent the corresponding class well.
To protect the knowledge involved in the classifiers, we update the classifiers more smoothly as follows:
\begin{equation}
\begin{aligned}
\mathbf{W}^i_t=\sum_{k=1}^{\hat{K}} \mathbf{\hat{W}}^k_t \cdot \frac{e^{\mathbf{p}_i^k}}{\sum_{j=1}^{\hat{K}} e^{\mathbf{p}_i^j}},
\end{aligned}
\label{update_the_target_classifiers}
\end{equation}
where $\mathbf{W}_t^i$ is the parameters for the $i^{th}$ target-domain identity in the next epoch, $\mathbf{p}_i ={\rm C}(\mathbf{c}_i|\mathbf{\hat{W}}_t)$ is the prediction of class center $\mathbf{c}_i$ with the parameters $\mathbf{\hat{W}}_t$ from the last epoch which includes $\hat{K}$ classes.
Note that the momentum for SGD is updated following the parameters in the process.

\subsection{Identity Learning}

The core of person re-identification is identifying the persons.
For two heterogeneous networks learning to extract discriminative representation, there are two level objective functions are applied.
Firstly, at the feature level, the triplet loss:
\begin{equation}
\begin{aligned}
\mathcal{L}_{tri}(\mathbf{f})=\frac{1}{n}\sum_{i=1}^n \max \{ \rho + d_p - d_n, 0 \},
\end{aligned}
\label{loss_triplet}
\end{equation}
is applied to guarantee the features can well represent their corresponding samples.
Where $\mathbf{f}$ represents a batch of the features, $n=|\mathbf{f}|$ is the size of the batch, 
$\rho$ is the tiniest margin between the distance to the furthest positive instance $d_p$ and the distance to the nearest negative instance $d_n$.
The relationship between two instances from the source domain depends on the ground-truth labels and the pseudo labels for target-domain samples.
Due to the different dimensions of the features extracted by heterogeneous networks, the triplet loss can only be applied in a certain subspace.

Then, in the logits level, we apply the cross-entropy loss with classifiers:
\begin{equation}
\begin{aligned}
\mathcal{L}_{Ttid}=-\frac{1}{n}\sum_{i=1}^n\log{P(\hat{\mathbf{y}}_i|{\rm C}(\mathbf{t}_i^T | [\mathbf{W}_s^T,\mathbf{W}_t^T]))},
\end{aligned}
\label{loss_teacher_learn_target_identity}
\end{equation}
\begin{equation}
\begin{aligned}
\mathcal{L}_{Stid}=-\frac{1}{n}\sum_{i=1}^n\log{P(\hat{\mathbf{y}}_i|{\rm C}(\mathbf{t}_i^S | \mathbf{W}_t^S))},
\end{aligned}
\label{loss_student_learn_target_identity}
\end{equation}
where $\hat{\mathbf{y}}_i$ is the pseudo label for target-domain example $\mathbf{x}_t^i$.
The trainable parameters $\mathbf{W}_s^T$, $\mathbf{W}_t^T$ and $\mathbf{W}_t^S$ respectively denote the classifier parameters for the teacher classifying source-domain samples, the teacher classifying target-domain samples, and the student classifying target-domain samples.
Meanwhile, to take advantage of the ground-truth label, the teacher also learns the source-domain knowledge by:
\begin{equation}
\begin{aligned}
\mathcal{L}_{Tsid}=-\frac{1}{n}\sum_{i=1}^n\log{P(\mathbf{y}_i|{\rm C}(\mathbf{s}_i^T | [\mathbf{W}_s^T,\mathbf{W}_t^T]))},
\end{aligned}
\label{loss_teacher_learn_source_identity}
\end{equation}
here, $\mathbf{y}_i$ is the ground-truth label for source-domain sample $\mathbf{x}_s^i$.
Note that, with Eq.(\ref{loss_teacher_learn_target_identity}) and Eq.(\ref{loss_teacher_learn_source_identity}), the classifier ${\rm C}(\mathbf{t}_i^T | [\mathbf{W}_s^T,\mathbf{W}_t^T])$ in the teacher has learned both two domain knowledge while classifier ${\rm C}(\mathbf{t}_i^T | \mathbf{W}_t^S)$ for the student learns the target-domain knowledge only.

\subsection{Asymmetric Mutual Learning (AML)}

Compare the structure of the Convolutional Neural Network (CNN) and Vision Transformer (ViT), the most evident difference is that the ViT can capture long-range information with its cascaded self-attention modules. 
However, CNN only focuses on the local limited by the size of the convolution kernel.
In addition, the CNN inductive bias, which includes assumptions of the data, can involve information that ViT may not consider and the convolution kernel with a deterministic shape guarantees spatial information.
More intuitively, the features extracted by the two networks have different dimensions.
It ensures the subspaces learned by the heterogeneous networks are different but makes feature-level constraints unusable.
The asymmetric distillation benefit from the difference in the patterns that two heterogeneous networks predict the identity.
And focus on twofold: to allow students access to knowledge from the different subspaces and transfer the knowledge from the source to the target.

%
To make the student learn from different subspaces and take advantage of the reliable source-domain labels, the proposed DAML transfers the identity knowledge from the teacher by reducing the Kullback-Leibler divergence between the predictions of the target-domain features as:
\begin{equation}
\begin{aligned}
\mathcal{L}_{id}=\frac{1}{n}\sum_{i=1}^n&{\rm C}(\mathbf{t}^T_i|\mathbf{W}_t^T)\log{\frac{{\rm C}(\mathbf{t}^S_i|\mathbf{W}_t^S)}{{\rm C}(\mathbf{t}^T_i|\mathbf{W}_t^T)}},
\end{aligned}
\label{equation_knwoledge_distillation}
\end{equation}
with the above objective function, the student can learn the knowledge from the teacher which adopts knowledge from both source and target domain with Eq.(\ref{loss_teacher_learn_target_identity}) and Eq.(\ref{loss_teacher_learn_source_identity}).
However, domain knowledge is also transferred to students and may harm the performance in the target domain.
The ideal way to alleviate the distribution effect is to make the teacher predict identities under the target domain.
Limited by the domain gap, the knowledge learned from the source domain can not be directly applied to the target domain.
And the goal of the proposed asymmetric mutual learning is to gain a student network that adapts to the target domain while benefiting from the source-domain identity knowledge.
Making source-domain predictions from the teacher similar to the student will transfer the classifying knowledge learned from the source domain to the target domain:
\begin{equation}
\begin{aligned}
\mathcal{L}_{dom}=\frac{1}{n}\sum_{i=1}^n&{\rm C}(\mathbf{s}^S_i|\mathbf{W}_t^S)\log{\frac{{\rm C}(\mathbf{s}^T_i|\mathbf{W}_t^T)}{{\rm C}(\mathbf{s}^S_i|\mathbf{W}_t^S)}},
\end{aligned}
\label{equation_eliminate_domain_gap}
\end{equation}
here, $\mathbf{W}_t^T$ learns source-domain knowledge with Eq.(\ref{loss_teacher_learn_source_identity}) while $\mathbf{W}_t^S$ only learns the knowledge from target domain.
%
%
Eq.(\ref{equation_eliminate_domain_gap}) focuses on making the teacher predict source-domain samples in the same way as the student.
In this way, the student can better adopt identity knowledge from the source domain without a domain gap as much as possible.
Compared with Eq.(\ref{equation_knwoledge_distillation}), the above equation distills the knowledge in a different direction and together make the student model can distinguish pedestrian in the target domain. 
%

\subsection{Optimization}
The total loss $\mathcal{L}$ of DAML can be summarized as:
\begin{equation}
\begin{aligned}
\mathcal{L}=&\big(\mathcal{L}_{Ttid} + \mathcal{L}_{tri}(\mathbf{t}^T)\big) +
             \big(\mathcal{L}_{Stid} + \mathcal{L}_{tri}(\mathbf{t}^S)\big) \\
            &\lambda_1\big(\mathcal{L}_{Tsid} + \mathcal{L}_{tri}(\mathbf{s}^T)\big)
            +\lambda_2\mathcal{L}_{id} + \lambda_3\mathcal{L}_{dom}
\end{aligned}
\label{full_loss_functions}
\end{equation}
where $\lambda_1$, $\lambda_2$, and $\lambda_3$ are hype-parameters to balance the contributions of individual loss terms.

\section{Experiments}

\subsection{Datasets}

\textbf{Datasets}
We evaluated our method on three public datasets \textbf{Market-1501}~\cite{conf/iccv/ZhengSTWWT15}, \textbf{CUHK-SYSU}~\cite{journals/corr/XiaoLWLW16} and \textbf{MSMT17}~\cite{conf/cvpr/WeiZ0018}.

\begin{itemize}
    \item \textbf{Market-1501} contains $32,668$ labeled images captured from $1,501$ identities by $6$ cameras.
    The training set has $12,936$ images of $751$ identities. 
    In addition, $3,368$ query images and $19,732$ gallery images from the other $750$ identities are used as the testing set.
    \item \textbf{CUHK-SYSU} includes $33,901$ labeled images of $8,432$ identities taken in diverse scenes.
    The training set is constructed with $5,532$ identities having $15,088$ images, and the rest is used for testing.
    There are $2,900$ images for the query and $6,978$ images for the gallery in the testing set.
    \item \textbf{MSMT17} is a large-scale dataset consisting of $126,441$ bounding boxes of $4,101$ identities caught on $12$ outdoor and $3$ indoor cameras.
    Among them, $32,621$ images of $1,041$ identities are used for training and $93,820$ of $3,060$ identities are used for testing.
\end{itemize}

\begin{table*}[t]

\caption{
Comparison of CMC (\%) and \emph{m}AP (\%) performances with the SOTA methods on \textbf{Market-1501}, \textbf{CUHK-SYSU} and \textbf{MSMT17}.
}

\centering
\resizebox{1.0\linewidth}{!}{
\setlength{\tabcolsep}{2mm}
\begin{tabular}{l||c|ccc||c|ccc}
\hline
\multirow{2}{*}{Method}  & \multicolumn{4}{c||}{Market-1501 $\rightarrow$ CUHK-SYSU} & \multicolumn{4}{c}{CUHK-SYSU $\rightarrow$ Market-1501}  \\
\cline{2-9} & \emph{m}AP & R1 & R5 & R10    & \emph{m}AP & R1 & R5 & R10   \\
\hline
Directly Transfer (IBN-ResNet-50)                   & 74.1  & 77.2  & 85.7  & 88.7  & 38.8  & 63.7  & 79.4  & 85.2 \\
Directly Transfer (ViT-Base)                        & 86.0  & 87.2  & 94.1  & 95.0  & 36.2  & 60.3  & 76.5  & 82.9 \\
\hline
UNRN~\cite{conf/aaai/ZhengLZZZ21}(AAAI'21)          & 62.3  & 64.1  & 76.9  & 82.0  & 70.9  & 86.7  & 92.8  & 94.5  \\
MMT(IBN-ResNet-50)~\cite{conf/iclr/GeCL20}(ICLR'20) & 78.4  & 81.0  & 89.7  & 92.2  & 76.0  & 88.8  & 95.2  & 97.0  \\
MEB-Net~\cite{conf/eccv/ZhaiYLJJ020}(ECCV'20)       & 81.1  & 83.2  & 90.9  & 93.1  & 69.3  & 84.0  & 92.9  & 95.2  \\
\hline
DAML (Ours)                                         & \textbf{84.3}  & \textbf{86.2}  & \textbf{92.6}  & \textbf{94.6}  & \textbf{84.1}  & \textbf{93.1}  & \textbf{97.7}  & \textbf{98.2} \\
\hline
Supervised (IBN-ResNet-50)                          & 90.8  & 95.2  & 96.6  & 89.0  & 83.0  & 94.1  & 97.4  & 98.4 \\
Supervised (ViT-Base)                               & 93.1  & 97.2  & 97.8  & 92.1  & 82.3  & 93.2  & 97.9  & 98.8 \\
\hline
\hline
\multirow{2}{*}{Method}  & \multicolumn{4}{c||}{Market-1501 $\rightarrow$ MSMT17} & \multicolumn{4}{c}{CUHK-SYSU $\rightarrow$ MSMT17}  \\
\cline{2-9} & \emph{m}AP & R1 & R5 & R10    & \emph{m}AP & R1 & R5 & R10   \\
\hline
Directly Transfer (IBN-ResNet-50)                   & 8.4   & 23.8  & 34.5  & 39.6  & 10.3  & 26.3  & 38.3  & 44.3 \\
Directly Transfer (ViT-Base)                        & 13.0  & 33.3  & 45.3  & 51.1  & 12.5  & 28.2  & 41.1  & 47.5 \\
\hline
MEB-Net~\cite{conf/eccv/ZhaiYLJJ020}(ECCV'20)       & 20.6  & 44.1  & 58.3  & 64.3  & 21.3  & 45.6  & 59.5  & 65.6  \\
UNRN~\cite{conf/aaai/ZhengLZZZ21}(AAAI'21)          & 25.3  & 52.4  & 64.7  & 69.7  & 12.6  & 31.1  & 43.8  & 49.7 \\
MMT(IBN-ResNet-50)~\cite{conf/iclr/GeCL20}(ICLR'20) & 26.6  & 54.4  & 67.6  & 72.9  & 24.0  & 49.0  & 63.0  & 68.6 \\
\hline
DAML (Ours)                                         & \textbf{41.4}  & \textbf{65.4}  & \textbf{76.0}  & \textbf{80.2}  & \textbf{44.0}  & \textbf{67.0}  & \textbf{78.0}  & \textbf{81.9}  \\
\hline
Supervised (ViT-Base)                               & 54.1  & 76.6  & 87.5  & 90.8  & 54.1  & 76.6  & 87.5  & 90.8 \\
Supervised (IBN-ResNet-50)                          & 49.9  & 79.2  & 88.2  & 91.3  & 49.9  & 79.2  & 88.2  & 91.3 \\
\hline
\end{tabular}
}

\label{table_comparison_SOTA}
\end{table*}

\begin{table}[!t]
\centering
\setlength{\tabcolsep}{1.0mm}
{
\caption{\small
Ablation study in terms of \emph{m}AP (\%) and CMC (\%) on \textbf{CUHK-SYSU (CS) $\rightarrow$ Market-1501 (M)}.
}
\label{table_ablation}
\resizebox{1.0\linewidth}{!}{
\renewcommand{\multirowsetup}{\centering}
\begin{tabular}{l||c|c}
\hline
\multirow{2}{*}{\ \ \ \ \ \ \ \ \ Method}  & \multicolumn{2}{c}{CS $\rightarrow$ M} \\
\cline{2-3} & \ \ \emph{m}AP \ \  & \ \ R1 \ \\
\hline
IBN-ResNet-50(Directly)                                         & 38.8 & 63.7 \\
ViT-Base(Directly)                                              & 36.2 & 60.3 \\
\hline
DAML w/o $\mathcal{L}_{Tsid} + \mathcal{L}_{tri}(\mathbf{s}^T)$ & 83.6 & 92.8 \\
DAML w/o $\mathcal{L}_{id}$                                     & 83.3 & 92.3 \\
DAML w/o $\mathcal{L}_{dom}$                                    & 83.6 & 92.5 \\
DAML w/o SCU                                                    & 81.0 & 91.0 \\
\hline
DAML                                                            & 84.1 & 93.1 \\
\hline
IBN-ResNet-50(Supervised)                                       & 83.0 & 94.1 \\
ViT-Base(Supervised)                                            & 82.3 & 93.2 \\
\hline
\end{tabular}
}
}
\end{table}

\subsection{Experiment Setting}

\textbf{Performance Metric:}
As a UDA task, we select one dataset as the source-domain dataset and another as the target-domain dataset.
The model is trained with the labeled source-domain training set and adapts the target domain through the unlabeled target-domain training set.
%
Then the performance is evaluated according to the student network which work on the target-domain testing set.
In our experiments, following the standard metrics, we employ the cumulative matching characteristic (CMC) curve and the mean average precision (\emph{m}AP) score.
Our experiments report rank-1, rank-5, and rank-10 accuracy and mAP scores.

\textbf{Implementation Details:} 
%
%
In the most common setting, we select IBN-ResNet-50~\cite{conf/eccv/PanLST18} as the teacher network and ViT-Base~\cite{conf/iclr/DosovitskiyB0WZ21} as the student network.
The batch size is set to $64$ for both source-domain and target-domain datasets. 
In one batch, the sampler will select $16$ identities and $4$ images for each identity according to the ground-truth label or pseudo label for two domains.
%
The input image has a fixed size of $256 \times 128$.

In the pre-training stage, we first train models $120$ epochs on the source-domain dataset.
%
The teacher CNN model is optimized by SGD with an initial learning rate of $1\times{10}^{-2}$ and weight decay of $5\times{10}^{-4}$ with a learning rate decays at $40^{th}$ and $70^{th}$ epoch.
The SGD optimizer is employed with a momentum of $0.9$ and the weight decay of $1\times{10}^{-4}$ for student ViT.
The learning rate is set to $8\times{10}^{-3}$, and the cosine schedule is applied.
The input images are augmented with random flip and randomly erase with $50\%$ probability.

In the fine-tuning stage, we adopted half the learning rate of the previous stage.
Specifically, the learning rate is set to $5\times10^{-3}$ for teacher CNN and $4\times10^{-3}$ for student ViT.
And the total number of training epochs is set to $60$.
The input images for two heterogeneous networks are randomly flipped and erased with $50\%$ probability.
%
When calculating the neighbors of an instance, the maximum acceptable distance $\alpha$ is $0.6$.
We generate the pseudo labels by DBSCAN~\cite{conf/kdd/EsterKSX96}.
For DBSCAN, we select $0.6$ as the maximum distance between neighbors and set the minimal number of neighbors to $2$ for \textbf{CUHK-SYSU} and $4$ for others.
The hype-parameters $\alpha$, $\lambda_1$, $\lambda_2$ and $\lambda_3$ are set to $0.5$, $0.1$, $0.7$ and $1.2$, respectively.
At the feature level, the margin $\rho$ for triplet loss is set to $1.2$.

\subsection{Comparison with State-of-the-art Methods}
Since Duke University terminated the \textbf{DukeMTMC}~\cite{conf/eccv/RistaniSZCT16} dataset, which has been widely used for evaluation of unsupervised domain adaptation person Re-ID task, the comparison becomes difficult.
To meet the moral and ethical requirements and provide a new baseline for comparison, we evaluate the performance of some representative works which have official open-source codes based on the \textbf{CUHK-SYSU} dataset.
%
And the results in \textbf{Market-1501 $\rightarrow$ MSMT17} setting is from the authors' reports.
We compare our DAML with state-of-the-art (SOTA) unsupervised domain adaptation person Re-ID approaches.
MMT~\cite{conf/iclr/GeCL20} applies two networks that have the same structure for learning from each other with both feature-level and logit-level constraints.
MEB-Net~\cite{conf/eccv/ZhaiYLJJ020} introduces three homogeneous networks, and the output of each network is considered comprehensively in the pseudo label generation.
Moreover, UNRN~\cite{conf/aaai/ZhengLZZZ21} designs a memory bank storing class centers from both source and target domains to mitigate the influence of noise labels.
As shown in Tab.~\ref{table_comparison_SOTA}, we evaluate the performance in four different manners, \textit{i.e.}, \textbf{Market-1501 $\rightarrow$ CUHK-SYSU}, \textbf{CUHK-SYSU $\rightarrow$ Market-1501}, \textbf{Market-1501 $\rightarrow$ MSMT17}, and \textbf{CUHK-SYSU $\rightarrow$ MSMT17}.

\textbf{Comparisons on large-scale datasets:}
The comparison results on \textbf{Market-1501 $\rightarrow$ MSMT17} and \textbf{CUHK-SYSU $\rightarrow$ MSMT17} are shown in the bottom of Table~\ref{table_comparison_SOTA}.
The proposed DAML outperforms existing SOTAs by large margins.
Specifically, DAML achieves the Rank-1 accuracy of $65.4\%$ and \emph{m}AP of $41.4\%$ in the \textbf{Market-1501 $\rightarrow$ MSMT17} setting, significantly improving the Rank-1 accuracy by $11.0\%$ and \emph{m}AP by $14.8\%$ over the SOTA MMT.
When compared to the SOTAs in \textbf{CUHK-SYSU $\rightarrow$ MSMT17} setting, the performance margin between our DAML and MMT is also significantly, e.g., the Rank-1 boost is $18.0\%$, and the \emph{m}AP boost is $20.0\%$.

\begin{table*}[!t]
\centering
\setlength{\tabcolsep}{1.0mm}
{
\caption{\small
Influence of different backbones in terms of \emph{m}AP (\%) and Rank-1 (\%) on \textbf{CUHK-SYSU (CS)} $\rightarrow$ \textbf{Market-1501 (M)}.
}
\resizebox{1.0\linewidth}{!}{
\renewcommand{\multirowsetup}{\centering}
\begin{tabular}{l|l|c|c||c|ccc}
\hline
\multirow{2}{*}{Method}  & \multirow{2}{*}{Backbone} & \multirow{2}{*}{\makecell{Training \\Parameter}} & \multirow{2}{*}{\makecell{Testing \\Parameter}} & \multicolumn{4}{c}{CS $\rightarrow$ M} \\
\cline{5-8} & & & & \ \emph{m}AP \ & \ R1 \ & \ R5 \ & \ R10 \ \\
\hline
MMT     & IBN-ResNet-50 + IBN-ResNet-50             & 99.8M     & 24.9M     & 76.0 & 88.8 & 95.2 & 97.0 \\
MMT     & ViT-Base + ViT-Base                       & 345.2M    & 86.3M     & 75.2 & 86.7 & 94.2 & 96.4 \\
UNRN     & ResNet-50-NL                             & 77.1M     & 38.5M     & 70.9 & 86.7 & 92.8 & 94.5 \\
UNRN     & ViT-Base                                 & 185.4M    & 86.3M     & 73.2 & 86.8 & 93.2 & 95.2 \\
\hline
DAML & IBN-ResNet-50 + ViT-Base                     & 111.2M    & 86.3M     & 84.1 & 93.1 & 97.7 & 98.2 \\
\hline
\end{tabular}
\label{table_parameter}
}

}
\end{table*}

\textbf{Comparisons on small-scale dataset:}
We also evaluate DAML on two small-scale target-domain datasets settings, \textbf{Market-1501 $\rightarrow$ CUHK-SYSU} and \textbf{CUHK-SYSU $\rightarrow$ Market-1501}, as shown in the top of Table~\ref{table_comparison_SOTA}.
Similar to the results on large-scale datasets, DAML consistently outperforms current SOTAs.
Specifically, we achieve Rank-1 accuracy of $86.2\%$ and \emph{m}AP of $84.3\%$ in \textbf{Market-1501 $\rightarrow$ CUHK-SYSU} setting.
Compared with the SOTA MEB-Net, the Rank-1 and \emph{m}AP respectively improved by $3.0\%$ and $3.2
\%$.
Meanwhile Rank-1 accuracy of $93.1\%$ and \emph{m}AP of $84.1\%$ are gained in \textbf{CUHK-SYSU $\rightarrow$ Market-1501} setting.
It improves the Rank-1 accuracy and \emph{m}AP by $4.3\%$ and $8.1\%$ compared with the SOTA MMT.
%
Note that the performance on \textbf{Market-1501 $\rightarrow$ CUHK-SYSU} setting is even worse than direct transfer. 
Because there are only two samples per class in \textbf{CUHK-SYSU} on average and it harms the pseudo label generation.
We will discuss this problem in section \textbf{Samples Augment}.

The above results demonstrate the outstanding performance of DAML thanks to its ability to learn knowledge from different subspaces and selectively transfer knowledge between two heterogeneous networks for unsupervised domain adaptation person Re-ID.

\subsection{Ablation Study}

In this section, we conduct ablation experiments on \textbf{CUHK-SYSU $\rightarrow$ Market-1501} setting to assess the contribution of each component by separately removing them from DAML for training and evaluation.
%
%
%

As shown in Table~\ref{table_ablation}, when removing $\mathcal{L}_{Tsid} + \mathcal{L}_{tri}(\mathbf{s}^T)$, the Rank-1 accuracy drops by $0.3\%$ and \emph{m}AP drops by $0.5\%$, since the reliable identity information is underutilized.
It illustrates that the ability to use the information in the source domain effectively is an essential factor in determining the model's performance.
It illustrates the essential to effectively use the information in the source domain
%
When removing $\mathcal{L}_{id}$, which helps student network to learn the identity knowledge from the teacher, the performance drops of Rank-1 and \emph{m}AP are $0.8\%$ and $0.8\%$, respectively, compared with the full DAML.
The performance drops due to the ignorance of the knowledge from the different subspaces in the logit level.
And the knowledge can still transfer to each other through the pseudo label generation.
%
%
Similarly, to validate the effectiveness of $\mathcal{L}_{dom}$, we remove it from DAML.
The result also shows the margin of the Rank-1 accuracy by $0.6\%$ and \emph{m}AP by $0.5\%$ to the complete DAML, which demonstrates that $\mathcal{L}_{dom}$ effectively helps to make the teacher network predict samples in 
%
The smooth classifier update (SCU) saves the knowledge learned in the last epoch.
When it is removed, the Rank-1 accuracy drops by $2.1\%$, and \emph{m}AP drops by $3.1\%$.
The results prove that learning the knowledge from different subspaces and taking advantage of correct identity information from the source domain are the two keys to solving UDA person Re-ID.

\subsection{Discussions}

\subsubsection{Influence of Backbone}
To meet the requirement of heterogeneous networks in the proposed DAML, we introduce the ViT-Base, which contains more trainable parameters as the backbone.
To clarify the source of performance growth, we repeat the experiments of MMT~\cite{conf/iclr/GeCL20} and UNRN~\cite{conf/aaai/ZhengLZZZ21} while replacing the backbone with ViT-Base.
As shown in Tab.~\ref{table_parameter}, "Backbone" represents the construction to extract features in the testing stage.
When the ViT-base replaces the backbone, the optimization process follows the setting in TransReID \cite{conf/iccv/He0WW0021}.
As shown in table.~\ref{table_parameter}, after replacing the backbone with ViT-Base, there is no significant change in results.
Limited by the symmetrical design in mutual learning manner and the high similarity in the classifying ways of ViTs, the Rank-1 and \emph{m}AP of MMT dropped $2.1\%$ and $0.8\%$, respectively.
For the UNRN method, which focuses on making pseudo labels reliable through the memory mechanism, the ViT brings $0.1\%$ and $2.3\%$ in Rank-1 and \emph{m}AP with much more parameters.
%
Based on the above experiments, we can conclude that the heterogeneous networks and the asymmetric learning strategy play a major role in the growth of performance.

\subsubsection{Heterogeneous Networks Analysis}
\begin{figure*}[t]
\centering
\includegraphics[width=2.0\columnwidth]{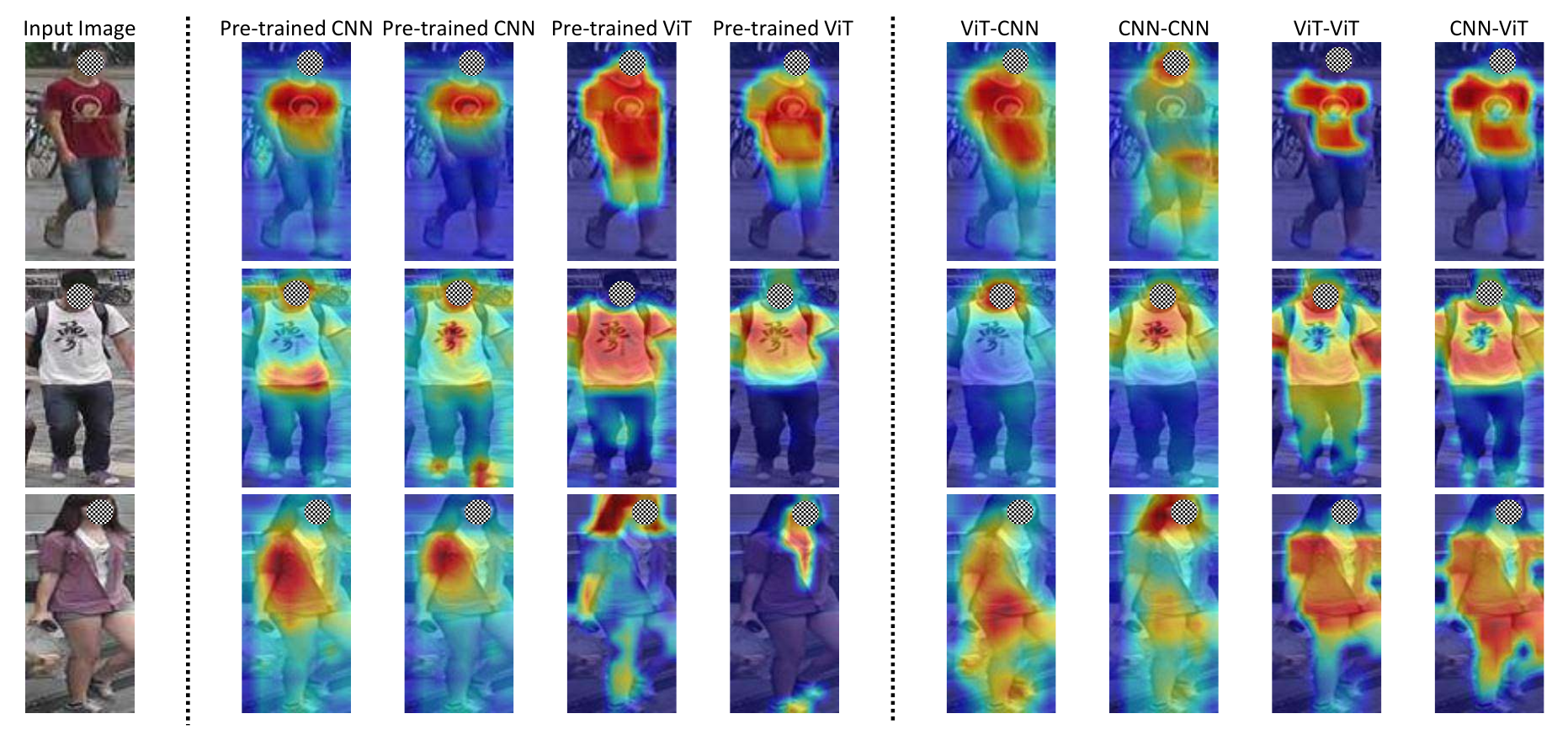}
\caption{\small
Visualization results of the models on \textbf{Market-1501}.
For each line, we show an input image, the area considered by the pre-trained ViT, the pre-trained CNN, and the different combinations of teacher and student in turn.
}
\label{fig3:CamGrad}
\end{figure*}
\begin{table}[!t]
\centering
\setlength{\tabcolsep}{1.0mm}
{
\caption{\small
Asymmetric distillation analysis in terms of \emph{m}AP (\%) and Rank-1 (\%) on \textbf{CUHK-SYSU (CS)} $\rightarrow$ \textbf{Market-1501 (M)}.
}
\resizebox{1.0\linewidth}{!}{
\renewcommand{\multirowsetup}{\centering}
\begin{tabular}{l|l||c|c}
\hline
\multicolumn{2}{c||}{Method} & \multicolumn{2}{c}{CS $\rightarrow$ M} \\
\hline
\ \ Teacher & \ \ Student & \ \emph{m}AP \ & \ \ R1 \   \\
\hline
IBN-ResNet-50   & ViT-Base          & 84.1 & 93.1 \\
ViT-Base        & ViT-Base          & 82.0 & 91.7 \\
ViT-Base        & IBN-ResNet-50     & 80.1 & 91.3 \\
IBN-ResNet-50   & IBN-ResNet-50     & 79.7 & 91.0 \\
\hline
\end{tabular}
\label{table_asymmetric}
}
}
\end{table}
One of the keys to improving UDA person Re-ID is learning knowledge from the different subspaces.
To illustrate the effect of heterogeneous networks, we train the proposed DAML with different combinations of teacher and student.
From Table.~\ref{table_asymmetric}, we can figure out that the student with a heterogeneous teacher will achieve better performance.
Specifically, the performance of student ViT improved by $0.7\%$ and $1.1\%$ in Rank-1 and \emph{m}AP with the asymmetric teacher. 
The results in the student CNN is similar, Rank-1 and \emph{m}AP are enhanced by $2.5\%$ and $5.3\%$.
These experimental results strongly prove the necessity of using two heterogeneous networks to work as the teacher and student.
And the knowledge from different subspaces has the capacity to help the student to learn broader knowledge.

When ViT is seen as the student, the benefit from the heterogeneous teacher is more evident than the improvement that CNN works as the student.
The heterogeneous teacher for ViT brings in $1.4\%$ and $2.1\%$ on Rank-1 and \emph{m}AP.
While it only improves Rank-1 and \emph{m}AP in $0.3\%$ and $0.4\%$ for the student CNN.
This phenomenon can be ascribed to the difference in the range of receptive field of these two networks that the former can consider the relationship between any two areas, but the size of convolution kernels limits the latter.
It gives ViT has the ability to learn the pattern that CNN applied to classify identities but not vice versa.

\subsubsection{Visualization}
The proposed DAML can make the student learn the knowledge from different subspaces.
To further illustrate the effectiveness of DAML, which can selectively transfer the knowledge between two networks, we apply Score-CAM \cite{conf/cvpr/WangWDYZDMH20} to visualize the pixel-wise attention areas on \textbf{CUHK-SYSU $\rightarrow$ Market-1501} setting.
Fig.~\ref{fig3:CamGrad} visualizes individual attention patterns for the three people from the target domain, where each column represents the attention area of the pre-trained CNN, ViT, and the different combinations of teacher-student.
From the first two columns, we can observe that the classifying patterns of CNN and ViT are different, which states the difference between their embedding spaces.
With these discrepancies, the heterogeneous networks can be improved by learning knowledge from heterogeneous networks.
In the last four columns, we can find that the networks mutual learning with heterogeneous networks can better consider the individual by the whole pedestrian while also taking into account many details that identify the persons more efficiently.
On the contrary, the recognition patterns of the networks that mutual learning with the same network have no significant change.
The visualization demonstrates the function of DAML in learning the knowledge from different subspaces improving the performance of the student.
%



\subsubsection{Samples Augment}
\begin{table}[!t]
\centering
\setlength{\tabcolsep}{1.0mm}
{
\caption{\small
Influence of Sample Augment for clustering algorithm in terms of \emph{m}AP (\%) and CMC (\%) on \textbf{Market-1501 (M) $\rightarrow$ CUHK-SYSU (CS)}.
}
\label{table_augment}
\resizebox{1.0\linewidth}{!}{
\renewcommand{\multirowsetup}{\centering}
\begin{tabular}{l|c||c|ccc}
\hline
\multirow{2}{*}{\ Method} & \multirow{2}{*}{\makecell{Repeat \\Times}} & \multicolumn{4}{c}{M $\rightarrow$ CS} \\
\cline{3-6} & & \emph{m}AP  & \ R1 \ & R5 & R10 \\
\hline
Directly Transfer                                           & - & 86.0 & 87.2 & 94.1 & 95.0 \\
\hline
DAML                                                        & 0 & 84.3 & 86.2 & 92.6 & 94.6 \\
\hline
\multirow{2}{*}{+ Random Crop}                              & 1 & 89.3 & 90.6 & 95.5 & 96.9 \\
                                                            & 2 & 89.1 & 90.4 & 95.4 & 96.6 \\
\hline
\multirow{2}{*}{+ Random Erase}                             & 1 & 88.3 & 90.0 & 95.0 & 96.3 \\
                                                            & 2 & 89.2 & 90.6 & 95.5 & 96.6 \\
\hline
\multirow{2}{*}{\makecell{+ Random Crop \\+ Random Erase}}  & 1 & 88.5 & 89.8 & 95.4 & 96.8 \\
                                                            & 2 & 87.6 & 89.0 & 95.2 & 96.6 \\
\hline
Supervised                                                  & - & 90.8 & 95.2 & 96.6 & 89.0 \\
\hline
\end{tabular}
}

}
\end{table}
Due to the small number of samples for each class in \textbf{CUHK-SYSU}, the performance of clustering algorithm is severely limited
%
%
The simplest and most direct way to address this problem is by augmenting the samples with random erase~\cite{conf/aaai/Zhong0KL020} and random crop, which can generate new samples while keeping the original identity when extracting features for the clustering algorithm.
As shown in Tab.~\ref{table_augment}, the performance with augmented data is better than the original. 
Specifically, the Rank-1 and \emph{m}AP are enhanced by $4.4\%$ and $5.0\%$ when applying the random crop method.
Similarly, the random erase improves Rank-1 and \emph{m}AP by $4.4\%$ and $4.9\%$.
The above experiment results state the necessity of enough samples for each class in the clustering algorithm.
Nevertheless, applying both random crop and random erase is not as effective as applying only one. 
The Rank-1 and \emph{m}AP are only increased by $3.6\%$ and $4.2\%$.
It suggests that the excessive augment method may harm identity knowledge and reduce the benefits from the augmented samples.
Compared to datasets collected for research, the number of identities and the number of samples in each identity are unknown in a real-world system.
And this information cannot be counted on the raw data unless annotated on them.
However, one of the advantages of unsupervised domain adaptation Re-ID is avoiding the annotation on the target domain, and it means the class-dependent super parameters are not available.
Because clustering algorithms elapse a long time to run on large datasets and take up most of the total training time, the super parameter selection experiments may be unacceptable for real-world systems.
Thus, a method without any clustering algorithm that still can mine identity knowledge from the target domain may make more sense for applying to the real world.
%
%
On the other hand, an efficient data augment method can restore the UDA methods based on clustering algorithm.

\section{Conclusion}
In this paper, we proposed the Dual-level Asymmetric Mutual Learning, termed DAML, to learn knowledge from a broader scope via asymmetric mutual learning with heterogeneous networks for unsupervised domain adaptation person Re-ID.
Our method aims to learn the knowledge from various subspaces and transfer the identity knowledge from the source to the target domain.
The former can improve feature expressiveness while also rectifying potential faults during training.
The latter takes full advantage of the identity knowledge from the source domain to improve the performance in the target domain.
Specifically, DAML first generates the pseudo labels according to the features extracted by heterogeneous networks, which are more reliable due to the consideration of various subspaces.
And the knowledge from two subspaces can be exchanged in a hard distillation manner in this process.
With the smooth classifier update, the classifiers can maintain the knowledge from the last epoch.
Then, the teacher will train on both source-domain and target-domain datasets to utilize the ground-truth label and transfer the knowledge to the target domain with the domain knowledge from the student.
To better adapt to the target domain, the student only trained on the target-domain dataset and benefited from the guidance from the teacher, which had learned the source-domain knowledge.
%
%
%
Experiments on four different experiment settings prove essential to learn the knowledge from various subspaces and demonstrate the effectiveness of the proposed DAML for unsupervised domain adaptation person Re-ID.

\bibliographystyle{IEEEtran}
\bibliography{tip2022}

\newpage

 




\vfill

\end{document}